%% file: main.tex
\documentclass[runningheads]{template_llncs}
\usepackage[T1]{fontenc}
\usepackage{graphicx}
\usepackage{algorithm}                 %
\usepackage[noend]{algpseudocode}      %
\makeatletter                          %
\def\BState{\State\hskip-\ALG@thistlm} %
\makeatother                           %

\usepackage{amsmath}  %
\usepackage{amssymb}  %
\usepackage{cases}    %
\usepackage{nicefrac} %
\usepackage{pifont}   %
\usepackage{array}    %
\usepackage{multirow} %
\usepackage{mdframed}  %
\global\mdfdefinestyle{mymdstyle}{linewidth=1pt}
\usepackage{enumitem}  %
\usepackage{lipsum}    %
\usepackage{wrapfig,lipsum,booktabs} %
\usepackage{multirow} %
\begin{document}

\title{Graph Reinforcement Learning for Operator Selection in the ALNS Metaheuristic}
\titlerunning{Graph RL for Operator Selection in ALNS}
\author{Syu-Ning Johnn*\inst{1} \and
Victor-Alexandru Darvariu*\inst{2,3} \and \\
Julia Handl\inst{4} \and
Joerg Kalcsics\inst{1} }
\authorrunning{S. Johnn et al.}
\institute{University of Edinburgh \\
\email{shunee.johnn@sms.ed.ac.uk; joerg.kalcsics@ed.ac.uk} \\
\and University College London \email{v.darvariu@cs.ucl.ac.uk} \and The Alan Turing Institute \\
\and University of Manchester \email{Julia.Handl@manchester.ac.uk} }

\maketitle 
\begin{abstract}
ALNS is a popular metaheuristic with renowned efficiency in solving combinatorial optimisation problems. However, despite 16 years of intensive research into ALNS, whether the embedded adaptive layer can efficiently select operators to improve the incumbent remains an open question.
In this work, we formulate the choice of operators as a Markov Decision Process, and propose a practical approach based on Deep Reinforcement Learning and Graph Neural Networks.  The results show that our proposed method achieves better performance than the classic ALNS adaptive layer due to the choice of operator being conditioned on the current solution.
We also discuss important considerations such as the size of the operator portfolio and the impact of the choice of operator scales. Notably, our approach can also save significant time and labour costs for handcrafting problem-specific operator portfolios.

\keywords{Adaptive Large Neighbourhood Search \and Markov Decision Process \and Deep Reinforcement Learning \and Graph Neural Networks}
\end{abstract}

\input{sections/1_intro.tex}

\input{sections/2_litreview.tex}

\input{sections/3_methods.tex}

\input{sections/4_experiments.tex}

\input{sections/5_conclusion.tex}

\bibliographystyle{splncs04}
\bibliography{bibliography} 
\end{document}

%% file: sections/1_intro.tex
\section{Introduction}
Adaptive large neighbourhood search (ALNS) is a metaheuristic introduced by Ropke and Pisinger~\cite{Ropke2006adaptive_abbreviated} to solve combinatorial optimisation problems (COPs) that iteratively deconstructs and reconstructs a part of the solution in the search for more promising solutions. This ``relax-and-reoptimise" process is executed via a pair of destroy and repair heuristics called operators. Based on the principle of Shaw's large neighbourhood search (LNS)~\cite{Shaw1999_abbreviated}, ALNS contains multiple operators and an adaptive layer that iteratively selects and applies different operator pairs from a predefined operator portfolio. This is typically an embedded Roulette Wheel (RW) algorithm that selects operators in a probabilistic fashion. %

ALNS is renowned for its efficiency in finding good-quality solutions within reasonable computational time. However, despite the wide use of ALNS for solving various COPs, the ways in which each ALNS component contributes to its general performance is not well understood. A recent ALNS state-of-the-art review~\cite{Mara2022survey_abbreviated} indicated that only 2 out of 252 papers go beyond the straightforward implementation and concentrate on component-based analysis, including~\cite{Santini2018comparison_abbreviated} which focuses on the selection of the ALNS acceptance criterion, and~\cite{Turkes2021meta_abbreviated} on the effectiveness of the ALNS adaptive layer for operator selection. 

We summarise two main deficiencies that exist in the current ALNS framework. 
Firstly, studies have shown that the adaptive layer has limited capability to dynamically select the best operators, despite being engineered to do so. Turke{\v{s}} et al.~\cite{Turkes2021meta_abbreviated} reported a mere 0.14\% average improvement brought by the adaptive layer from the analysis of 25 ALNS implementations, indicating a need for a more efficient operator selection mechanism that reflects the contribution of individual operators accurately.
Secondly, operator portfolio design for a particular problem can require considerable manual evaluation~\cite{Mara2022survey_abbreviated}. The choice of portfolio size is also delicate: too few operators might not enable the search to visit unexplored neighbourhoods, but a plethora of operators can introduce noise to the adaptive layer.
To mitigate these deficiencies, we make the following contributions:
\begin{itemize}[label=\textbullet, leftmargin=3mm]
    \item We formulate the choice of a sequence of operators as a Markov Decision Process (MDP), in which an agent receives a reward proportional to the improvement in the solution. We draw a correspondence between value-based Reinforcement Learning (RL) methods used to solve MDPs, such as Q-learning, and the classic RW update used in ALNS. A key insight is that RL estimates are conditioned on the current solution, while RW updates are independent of it, which indicates the potential to learn a stronger operator selector through the RL framework;
    \item We propose a practical approach based on Deep RL for learning to select operators. Furthermore, we highlight the potential of Graph Neural Networks (GNNs) for generalizing to larger problem instances than seen during training;
    \item We carry out an extensive evaluation that includes a large selection of representative operators from the literature. Our results demonstrate that the proposed approach performs significantly better than the RW mechanism. We also analyse the impact of important practical considerations such as portfolio sizes and destroy operator scales on the optimality of the solutions.
\end{itemize}

%% file: sections/2_litreview.tex
\section{Literature Review}

In the last decade, training Machine Learning (ML) methods to solve highly complex COPs has become increasingly prominent \cite{Bengio2021machine_abbreviated}, especially for the Vehicle Routing Problem (VRP) and its variants \cite{Bai2021analytics_abbreviated}. 
Several pioneering studies applied RL to directly construct solutions for routing-related problems. Bello et al. \cite{Bello2016neural} used policy gradient algorithms to tackle the Travelling Salesperson Problem (TSP). Nazari et al. \cite{Nazari2018reinforcement_abbreviated} proposed an end-to-end framework that outputs solutions directly from the routing-based problem instances. Moreover, Kool et al. \cite{Kool2018attention} proposed a construction heuristic that consists of an attention-based decoder trained with RL to regressively build solutions for the TSP and its variants. 

ML can also be applied in many cases to enhance existing solution approaches, especially in the field of metaheuristics %
\cite{Talbi2021machine}. The reader is referred to the work of Karimi-Mamaghan et al.~\cite{Karimi2022machine_abbreviated} for a comprehensive review on the integration of ML and metaheuristics to tackle COPs. 

Several recent studies focused on integrating ML with classic LNS, which can be viewed as a simplified version of the ALNS metaheuristic without the adaptive layer for operator selection. 
As the first paper on this topic, Hottung and Tierney~\cite{Hottung2019neural} proposed $2$ generalised random-based destroy operators and a single repair operator with automated learning based on a deep neural network with an attention mechanism. Their work was the first to consider the application of RL to LNS for solving a VRP, and achieved solutions of better quality than classic optimisation approaches. Nevertheless, their proposed learning mechanism only focuses on repairing incomplete solutions during the repair phase.
In another work, Falkner et al.~\cite{Falkner2022large} integrated a pre-trained neural construction heuristic as the repair operator in the LNS framework to solve the VRP with time windows. The destroy procedures remain handcrafted and are classified into $2$ groups without any learning involved. 
Moreover, Oberweger et al.~\cite{Oberweger2022learning_abbreviated} enhanced the LNS framework with an ML-guided destroy operator to solve a staff rostering problem. For the reconstruction phase, the authors developed a mixed-integer linear
program as a repair method.
Lastly, Syed et al.~\cite{Syed2019neural} proposed a neural network in an LNS setting to solve a vehicle ride-hailing problem. However, it uses supervised learning, which requires a large training dataset and, furthermore, can only perform as well as the algorithm used for its generation.

A very recent concurrent work by Reijnen et al.~\cite{reijnen2022operator} also applies Deep RL to improve ALNS operator selection. It considers a state space that only uses information about the search status (such as the search step), ignoring information about the current solution. In contrast to this, the design of our approach focuses on isolating the problem of operator selection from the search process, and proposing a learning mechanism that is conditioned on the decision space characteristics of the current solution. Furthermore, a fixed operator portfolio consisting of $4$ destroy and $3$ repair operators is used in their evaluation. In contrast, we propose a more robust operator selection system compatible with various operator portfolios of different sizes and train the system independently prior to integration with ALNS. Our approach also proposes the use of GNNs for scaling to large instances.

%% file: sections/3_methods.tex
\section{Methodology}\label{sec3}
\subsection{Classic ALNS Algorithm}\label{subsec3.1}
In ALNS \cite{Ropke2006adaptive_abbreviated}, an initial solution is relaxed and re-optimised through iteratively employing a pair comprising a destroy operator $o_i^-\in \mathcal{D}$ and a repair operator $o_i^+\in \mathcal{R}$ to form the new incumbent. 
The destroy scale $d$, which is randomly drawn or set as a hyper-parameter,  describes the proportion of the solution that is destructed and reconstructed.
In ALNS, the search can be divided into sequential segments, during which an initial score $\psi_i=0$ is assigned to each operator (indexed by $i$) at the beginning and is increased by $\delta$ each time a new incumbent is formed using an operator pair that includes $i$. Depending on the incumbent quality, the score is increased by $\delta_1$ if the newly-found incumbent is a global best solution, $\delta_2$ for a local best one, and $\delta_3$ for an accepted yet worse local solution, where $\delta_1>\delta_2>\delta_3$. At the end of each segment, the cumulated score for each operator $i$ and the number of times $N_i$ it was selected are used to compute a weight $w_i$ that estimates the operator's capability to find promising solutions. As shown in Equation \eqref{eq1_rw_update_formulation}, for each operator employed in the current segment $K$, its weight for the next segment $K+1$ is updated using a weighted average of the historical weight $w_{i,K}$ and its average performance in segment $K$. 
\begin{equation}
w_{i,K+1} = 
    \begin{cases}
    (1-\alpha_{\text{\tiny{RW}}})\cdot w_{i,K} + \alpha_{\text{\tiny{RW}}} \cdot\frac{\psi_i}{N_i} & \text{if}\ \psi_i > 0, \\
    w_{i,K} & \text{if}\ \psi_i = 0,
    \end{cases} \label{eq1_rw_update_formulation}
\end{equation}
For each iteration within the segment, a pair of operators is selected using the RW selection algorithm with probabilities $w_{i,K}^{-} \big/ \sum_{j\in \mathcal{D}} w_{j,K}^{-}$ and  $w_{i,K}^{+} \big/ \sum_{j\in \mathcal{R}} w_{j,K}^{+}$, where $w_{i,K}^{-/+}$ is the weight associated with each operator $i$ in any given segment $K$. Initially, all operators are assigned the same score and therefore have the same selection probability.
Once a new solution is formed, an ALNS acceptance mechanism, typically used in Simulated Annealing (SA), determines whether the newly-formed solution is accepted as the new incumbent to start the next iteration. The probabilistic acceptance mechanism helps to diversify the search and reduce the chance of becoming trapped in a non-promising local neighbourhood. The process continues until certain stopping criteria are met.

\subsection{Operator Selection as a Markov Decision Process}\label{subsec3.2}

\noindent \textbf{Blueprint of our Approach.} Our learning-based approach to improve the operator selection in ALNS consists, at a high level, of the following two steps. 
Firstly, we aim to isolate operator choice from the considerations of the SA process in ALNS, which introduces additional noise for navigating the solution space that may obscure the operators' contributions. To achieve this, we formulate operator selection for the COP as a standalone Markov Decision Process (MDP), in which an agent is given a limited budget of operators, and must learn to select those that lead to the best solutions. Secondly, the learned model is integrated into the ALNS loop and used to select operators in the SA process.

\vspace{2mm}
\noindent \textbf{MDP Fundamentals.} An MDP is a tuple $(\mathcal{S}, \mathcal{A}, P, R)$. In each state $s\in \mathcal{S}$, the agent selects an action $a\in \mathcal{A}(s)$ out of a set of valid actions, receiving a reward $r$ according to a reward function $R(s,a)$. Afterwards, the agent transitions to a new state $s'$ that depends on $P(s' | s, a)$, which is the transition function that governs the environment dynamics. 
Interactions happen in episodes, each of which is a finite sequence of $(s, a, r, s')$ pairs, until a terminal state is reached. Actions are selected by the agent through the \textit{policy} $\pi\left(a|s\right)$ that completely specifies its behaviour. The state-action value function $Q(s, a)$ is the expected reward the agent receives by picking action $a$ at a given state $s$, then following $\pi$.

\begin{figure}[htp]   \centering\includegraphics[width=\textwidth]{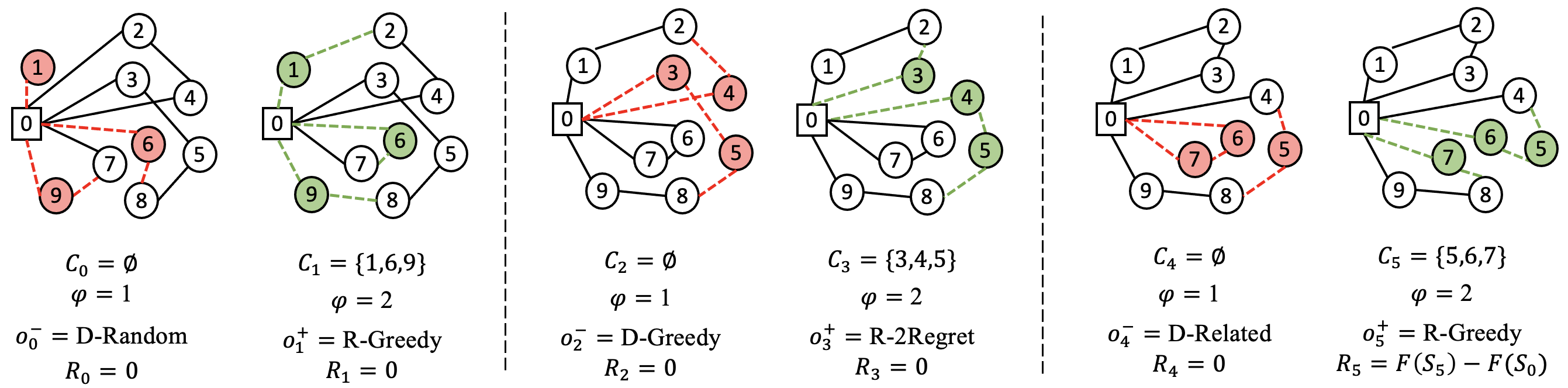}
    \caption{Illustration of an MDP episode with budget $b=3$ and destroy scale $d=3$. The action spaces contain 3 destroy operators $\mathcal{D}=\{\text{Random, Greedy, Related}\}$ and 2 repairs $\mathcal{R}=\{\text{Greedy, 2Regret}\}$. The agent begins at state $S_0$ with $C_0=\emptyset$ and routes $J_0=\{[1], [2,4], [3,5,8,6], [7,9] \}$, selecting operators $o_0^- = \text{Random}$ and $o_1^+=\text{Greedy}$ to reach $S_2$. The episode continues until the budget is exhausted and the terminal state $S_5$ with routes $J_5 =\{ [1,2,3], [4,5,6], [7,8,9] \}$ is reached. Finally, it receives a reward proportional to the improvement in solution quality.
    }
    \label{fig-MDP-sample}
\end{figure}

\vspace{2mm}
\noindent \textbf{MDP Formulation.} We are given an undirected graph $G = (V, E)$ defined by the given COP and a feature matrix $\mathbf{X}$ in which each row contains information about the node such as coordinates, demand, and distance. %
We formulate the MDP as below. A visualisation of an episode is shown in Figure \ref{fig-MDP-sample}. 
\begin{itemize}[label=\textbullet, leftmargin=3mm]
    \item \textit{States $\mathcal{S}$}: each state $S_t$ is a tuple $(G, \mathbf{X}, J_t, C_t, \varphi, b_t)$, wherein the graph $G$ and feature matrix $\mathbf{X}$ remain static. $J_t$ is the set of \textit{tours} that start and end at the depot, forming the solution at time $t$. The removal list $C_t= V \setminus J_t$ holds all the $d$ nodes temporarily removed from the solution. $\varphi$ indicates the \textit{phase}: whether a destroy or repair operator is eligible to be applied. Finally, $b_t$ indicates the operator pair budget available to the agent.
    \item \textit{Actions $\mathcal{A}$} involve the selection of an operator $o_t$, with those available defined as $\mathcal{D}$ if $\varphi=1$ (i.e., we are in the destroy phase), and $\mathcal{R}$ otherwise.
    
    \item \textit{Transitions $P$} apply the selected operator $o_t$ to the current solution. Applying a destroy operator removes $d$ nodes from $J_t$ and places them in the removal list $C_t$. Using a repair operator reinserts the nodes from $C_t$ into $J_t$, leaving $C_t$ empty and the solution $J_t$ complete, and decreases the operator pair budget by $1$. Transitions are stochastic due to the inherent randomness of the operators.
    \item \textit{Rewards $R$} are provided once the operator budget is exhausted and the improvement in solution quality can be assessed via an objective function $F$. Concretely, $R(S_t, A_t) =  F(S_t) - F(S_0)$ if $b_t = 0$, and $0$ otherwise.

\end{itemize}
\vspace{-3mm}

\vspace{-4mm}
\subsection{Learning an Operator Selection Policy}\label{subsec3.3}

\noindent \textbf{Q-learning and relationship to Roulette Wheel update.} Q-learning~\cite{Watkins1992Qlearning} is a model-free RL approach for solving MDPs that relies on estimating the state-action value function $Q(s,a)$, from which a policy $\pi$ can be derived by acting greedily with respect to it. The agent's interactions with the environment generate $(s, a, r, s')$ tuples, and its estimates are updated according to the rule 
\begin{equation}
    Q(s, a) \leftarrow (1-\alpha_{\text{\tiny{RL}}})\cdot Q(s, a) + \alpha_{\text{\tiny{RL}}}\cdot \left(r + \gamma\cdot \max_{a' \in \mathcal{A}(s')} Q(s', a')\right) \label{eq3_Q_learning_formulation}
\end{equation}
where $\alpha_{\text{\tiny{RL}}}$ is the learning rate, and $\gamma$ trades immediate versus long-term rewards.
Written in this form, comparing the Q-learning update in Equation~\eqref{eq3_Q_learning_formulation} and the classic RW update in Equation~\eqref{eq1_rw_update_formulation}, we notice that both use a weighted factor to balance two terms representing the historical and current estimates of performance. The key difference is that the Q-learning update is conditioned on the state and hence captures more information that may be used to select a relevant operator, while the RW update simply averages the gains of the operators irrespective of the context in which they were applied. Therefore, RW can be interpreted as a very rough approximation of the Q-learning update and, intuitively, using information about the state can allow us to obtain operator selection policies that perform at least as well. 
This means that Q-learning requires higher sample complexity. However, this was not an issue in practice, as we found a relatively low number of training steps suffices to reach a good policy.

\vspace{2mm}
\noindent \textbf{Function Approximation and Graph Neural Networks.} In problems with large state spaces, neural networks are commonly used to perform \textit{function approximation} of the $Q(s,a)$ function. This helps to generalize between states that, while not being identical, share common characteristics and hence may lead to similar future rewards. The Deep Q-Network (DQN) algorithm~\cite{Mnih2015Human}, which uses this principle together with replay buffers and target networks, has been used for successfully approaching a variety of decision-making tasks.

In this work, we consider two possible neural network architectures. Firstly, we use a Multi-Layer Perceptron (MLP) formed of layers that apply a linear transformation of the inputs followed by a non-linear activation function. Despite their simplicity, MLPs are known to be universal function approximators. Secondly, we consider Graph Neural Network (GNN) architectures~\cite{Scarselli2008GNN}, that are explicitly designed to operate on graph-structured data. Such architectures compute an embedding for each node in the graph by iteratively aggregating the features of neighbouring nodes, resulting in node embeddings that encode both structural and feature-based information. A desirable characteristic of many GNN architectures is that their parametrization can be independent of the size of the input graph. Hence, they enable learning an approximation of the state-action value function on small instances and applying it directly on large instances -- an appealing approach for COPs~\cite{Bengio2021machine_abbreviated}.

\vspace{2mm}
\noindent \textbf{Integrating the model with ALNS.} As mentioned above, the resulting learned policy acts greedily with respect to the learned state-action value function, always choosing the action with the highest expected cumulative reward. This might prove problematic once integrated within ALNS, given that, in principle, greediness may cause the search to become trapped in local optima. To instead obtain a \textit{probabilistic} policy, we use a softmax function as shown in Equation~\eqref{eq4_softmax}, in which the temperature $\tau$ allows adjusting the level of greediness of the policy. Specifically, probabilities are uniform when $\tau\to\infty$, whereas the action with the highest expected reward has probability approaching $1$ when  $\tau \to 0$.
\begin{equation}
    \pi_{\tau}(a|s) = \frac{\text{exp}(Q(s,a)/\tau)}{\sum_{a' \in \mathcal{A}(s) } \text{exp}(Q(s, a')/\tau)} \label{eq4_softmax}
\end{equation}

\subsection{Operators for ALNS}\label{subsec3.4}
In the literature, operators are carefully tailored to fit different problem structures and features. Despite the large variety of operator designs, the mechanisms behind them are surprisingly similar to the first version of ALNS~\cite{Ropke2006adaptive_abbreviated}. We conducted a thorough analysis of operators in the literature and have identified the following 3 classes: 
\textit{random-based} destroy that randomly removes $d$ nodes according to specific availability criteria, 
\textit{greedy-based} destroy that removes the top-ranking $d$ nodes with respect to a particular measure, and 
\textit{related-based} destroy as an extension of Shaw's destroy \cite{Shaw1999_abbreviated} that removes the most similar $d$ nodes according to a certain proximity value. 
Variations can include perturbations or using problem-specific features including distance, time, cost, workload, demand level, inventory level, removal gain, historical information, etc. 

Barring a few \textit{random-based} operators, almost all current repair operator designs are related to \textit{greedy-based} mechanisms that insert each node at the position with the smallest cost. Variations can include a pre-sorting that changes the order of node insertions according to certain criteria, including global minimum insertion or smallest regret value. Others can have a noise factor that perturbs the insertion cost values, or use restrictions based on historical information.

%% file: sections/4_experiments.tex
\vspace{-2mm}
\section{Experiments}

\subsection{Experimental Setup}

\noindent \textbf{Problem Settings.} In this work, we consider the Capacitated Vehicle Routing Problem (CVRP) with a single depot, a group of customer nodes and a number of homogeneous vehicles each visiting an individual group of customer nodes. The capacity restriction applies to the total carrying load of vehicles. Each customer node can only be visited once. 
We use the R, C and RC instances (random, clustered, and mixed random-clustered nodes) of the Solomon dataset \cite{Solomon1987dataset} each containing a depot and $100$ customers. We assign the vehicle capacity to be $200$, and adjust it proportionally if scaling down the instance to fewer customers.

For the portfolio design, 
we identified 12 popular destroy operators from the ALNS literature that span the representative categories described in Section~\ref{subsec3.4}:
the random-based variations 
\textit{random node destroy} \cite{Ropke2006adaptive_abbreviated} and %
\textit{random route destroy} \cite{Demir2012adaptive_abbreviated}, %
the greedy-based variations 
\textit{worst-node removal} \cite{Ropke2006adaptive_abbreviated}, %
\textit{neighbourhood removal} \cite{Demir2012adaptive_abbreviated} and %
\textit{greedy route destroy} \cite{Keskin2016partial_abbreviated}, %
and the related-based variations 
\textit{proximity destroy} \cite{Demir2012adaptive_abbreviated}, %
\textit{cluster destroy} \cite{Pisinger2007general_abbreviated}, %
\textit{node neighbourhood destroy} \cite{Demir2012adaptive_abbreviated}, %
\textit{zone destroy} \cite{Emec2016adaptive_abbreviated}, %
\textit{route neighbourhood destroy} \cite{Emec2016adaptive_abbreviated}, %
\textit{pair destroy} \cite{Mancini2016real_abbreviated} and  %
\textit{historical node-pair removal} \cite{Pisinger2007general_abbreviated}. %
The repair operator portfolio is comparatively smaller. We include the group of classic \textit{greedy repair} \cite{Ropke2006adaptive_abbreviated} %
and \textit{k-regret repair} \cite{Pisinger2007general_abbreviated} for $k=2$. %

\vspace{2mm}
\noindent \textbf{Operator Selection Approaches.} The proposed DQN agent is compared to the following approaches. As a baseline, we consider a uniform Random sampling (RAN) of operators. We also compare against the classic RW (CRW), which can only be used within ALNS since it requires information about the SA outcomes and search progress. To make the RW mechanism applicable in the MDP setting, we make the following adaptations to obtain a method we call Learned RW (LRW). Firstly, in Equation~\eqref{eq1_rw_update_formulation}, we replace the manually-defined operator scores $\psi$ computed from the discretised $\delta$ with the continuous objective value $F$. We also adjust the reward feedback frequency from every operator pair in RW to every episode in the MDP. Preliminary experimental results suggested that the performance difference between the LRW and CRW is within 2\% when applied in ALNS without any prior training.

\vspace{2mm}
\noindent \textbf{Training and Evaluation Methodology.} For each instance, we generate $3$ distinct sets $\mathcal{J}^{\text{train}}, \mathcal{J}^{\text{validate}}, \mathcal{J}^{\text{test}}$ of $128$ randomly initialized tours each. $\mathcal{J}^{\text{train}}$ is used by DQN and LRW for model training. $\mathcal{J}^{\text{validate}}$ is used for hyperparameter tuning and model selection. Finally, $\mathcal{J}^{\text{test}}$ is used to perform the final evaluation and obtain the reported results. There are two evaluation ``modes'': MDP-compatible agents can be evaluated in a standalone fashion given an operator budget (CRW is excluded), while all operators (including CRW) can be evaluated on the end ALNS task. Training and evaluation is repeated across $10$ random seeds for all agents, which are used to compute confidence intervals.

\vspace{2mm}
\noindent \textbf{DQN Architectures and Inputs.} For the DQN, we consider MLP and GNN representations. The MLP has 256 units in the first hidden layer, with the subsequent layers having half the size. As a GNN, we opt for the GAT~\cite{Velivckovic2017GNN}, which allows for flexible aggregation of neighbour features. We use $3$ layers and a dimension of node embeddings equal to $32$. Both use a learning rate of $\alpha_\text{\tiny{RL}}=0.0005$ and are trained for $15 \cdot 10^3$ and $25 \cdot 10^3$ steps respectively. The DQN exploration rate $\epsilon$ is linearly decayed from $1$ to $0.1$ in the first 10\% of steps, then remains fixed. The replay buffer size is equal to 20\% of the number of steps. 
To obtain the inputs, we construct vectors $\tilde{\textbf{x}}^i_t$ that concatenate the static instance-specific features $\textbf{x}^i$ with time-dependant relevant information such as whether the node $i$ is routed in a tour and the number of tours in $J_t$. For the MLP, we stack the vectors in a matrix $\widetilde{\mathbf{X}}_t$ as inputs, while for the GNN the node features are provided directly. Unless otherwise stated, we use a softmax temperature $\tau = 0.01$. %

\subsection{Experimental Results}

\noindent \textbf{Evaluating Agents within MDP Framework.} In this experiment, we compare the cumulative rewards gained by the DQN with an MLP representation, LRW and RAN agents on the test set $\mathcal{J}^{\text{test}}$ after undergoing training. To make the training and evaluation processes less computationally intensive, we use the first 20 customer nodes and the depot from the Solomon R, C and RC instances. We define operator portfolios of different sizes ranging from 2 to 12 by sequentially adding the 12 destroy operators introduced above, together with the 2 repair operators. The destroy scale is fixed as $d=4$ and the operator pair budget is $b=10$, yielding MDP episodes of length $20$.

\begin{table}[htp]
\caption{MDP evaluation results: cumulative rewards gained by the DQN, RAN and LRW agents with destroy portfolios $\mathcal{D}$ of different sizes. Higher is better.} 
\label{table1_MDP_evaluation}
\centering
\resizebox{0.99\textwidth}{!}{
    \input{tables/mdp_evaluation.tex}
}
\end{table}
Table \ref{table1_MDP_evaluation} shows that the DQN agent is able to outperform competing methods as the size of $\mathcal{D}$ grows. 
When the destroy portfolio is smaller than $3$, the DQN agent performs slightly worse due to the limited action space in which the impact of the selected actions is difficult to distinguish from chance. The DQN agent also yields smaller confidence intervals and hence a steadier performance. As expected, the RAN agent fails to show a clear increase in rewards as the portfolio size grows. The LRW agent, although showing a certain improvement, performs significantly worse than the DQN. 
Two performance jumps in the DQN and LRW agents were observed: from size 4 to 5 and 11 to 12 for all $3$ instances, the reason for which is the inclusion of a more efficient operator in the portfolio that suits the behaviour of a greedy-based agent. 

\begin{table}[htp]
\caption{Evaluating operator selection approaches in ALNS with destroy portfolios $\mathcal{D}$ of different sizes. Values represent the average and best objective values found within a fixed number of iterations, using each approach to select operators. Lower is better.}
\label{table2_ALNS_evaluation_3instances}
\centering
\resizebox{0.99\textwidth}{!}{
    \input{tables/alns_evaluation.tex}
}
\end{table}

\vspace{2mm}
\noindent \textbf{ALNS Evaluation.} Using the same experimental setup as above, we apply the operator selection approaches within ALNS with a fixed number of iterations. As shown in Table~\ref{table2_ALNS_evaluation_3instances}, the DQN agent yields the lowest objective values (best results) when used to perform operator selection in ALNS for portfolios larger than $5$. Interestingly, LRW is able to perform substantially better than CRW due to having undergone training on a different dataset of solutions prior to being applied. Instead, the performance of CRW is indistinguishable from RAN in the setting across all the $3$ instances.

\begin{figure}[htp]
\centering
\includegraphics[width=\textwidth]{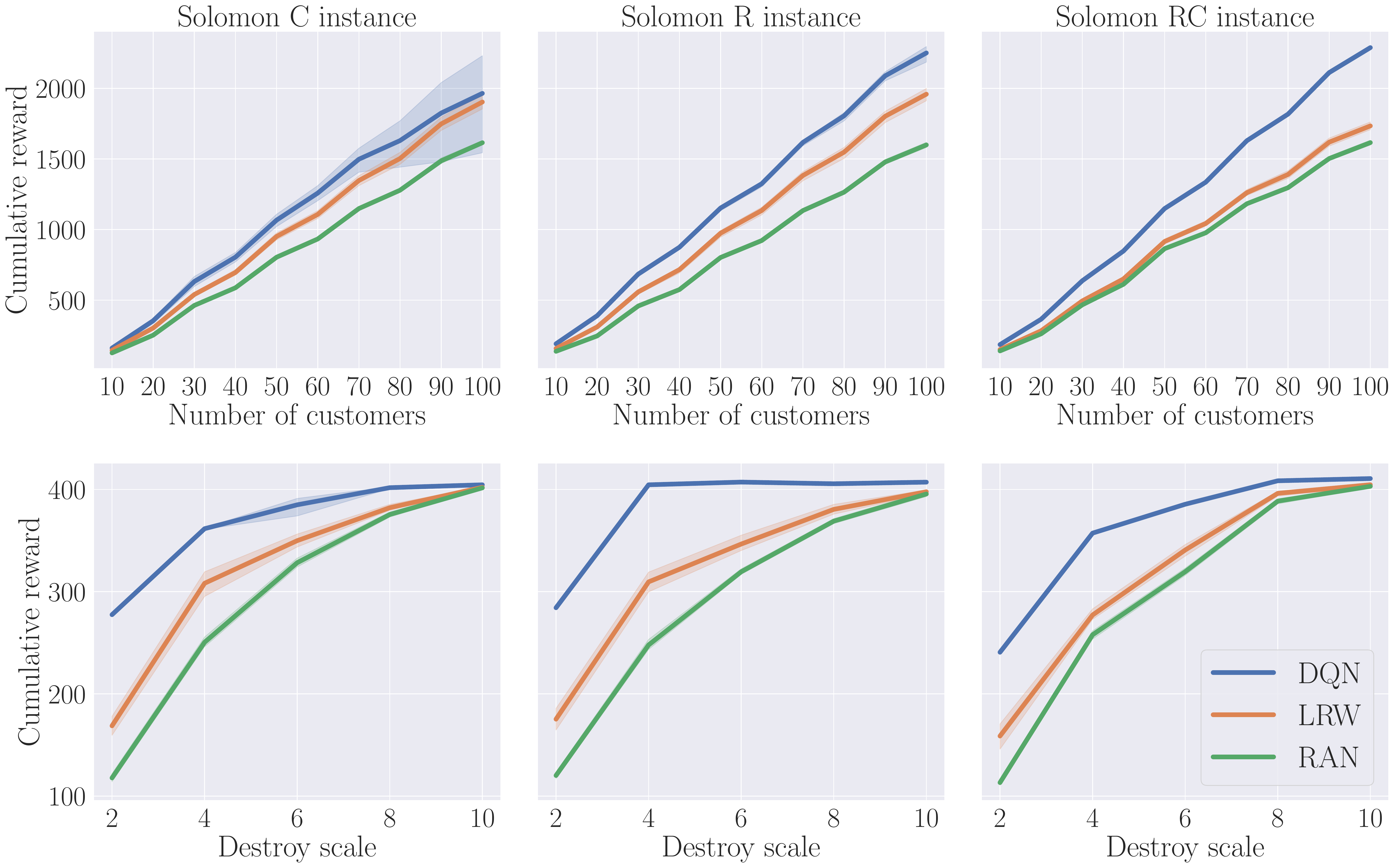}
\caption{Top: cumulative rewards for the DQN, LRW and RAN agents with GNN representation. Bottom: performance as a function of destroy scales. Higher is better.} 
\label{fig2:gnn_increase_scale}
\end{figure}

\vspace{2mm}
\noindent \textbf{Scaling to Larger Instances with GNN.} In this experiment, we train the DQN with a GNN representation and the LRW on instances of size $20$, then evaluate them in an MDP setting on instances of size up to $100$. The operator budget is kept the same while the destroy scale is increased proportionally to the size, i.e., $d=\nicefrac{n}{5}$. We use the largest destroy portfolio with $|\mathcal{D}|=12$. As shown in the top half of Figure~\ref{fig2:gnn_increase_scale}, the DQN+GNN agent outperforms the other methods, suggesting the strong generalization of the learned operator selection policies. A larger confidence interval is observed for the C instance, due to $1$ model seed that generalizes poorly on $\mathcal{J}^{\text{test}}$ despite good performance on $\mathcal{J}^{\text{validate}}$.

\vspace{2mm}
\noindent \textbf{Impact of Destroy Scale.} Furthermore, we analyse the impact of the destroy scale on the agents' performances in the MDP setting, with a smaller scale implying the removal and reinsertion of a smaller proportion of nodes. We vary the destroy scale $d \in [2,4,6,8,10]$ with destroy portfolio $|\mathcal{D}|=12$ on 20 nodes. Results are shown in the bottom half of Figure~\ref{fig2:gnn_increase_scale}. The gap between the DQN and other methods is largest for the smallest scale, suggesting that a careful selection of operators to remove the most expensive nodes contributes more significantly to better solution quality. In contrast, a larger destroy scale requires building up the solution from the ground, stressing the operators’ reconstruction ability rather than the operator selection policy. When increasing the destroy scale, the cumulative rewards gained by different agents all converge to a similar level. 

\begin{wrapfigure}{r}{0.33\textwidth}
\vspace{-2.85\baselineskip}
\begin{center}
\includegraphics[width=0.33\textwidth]{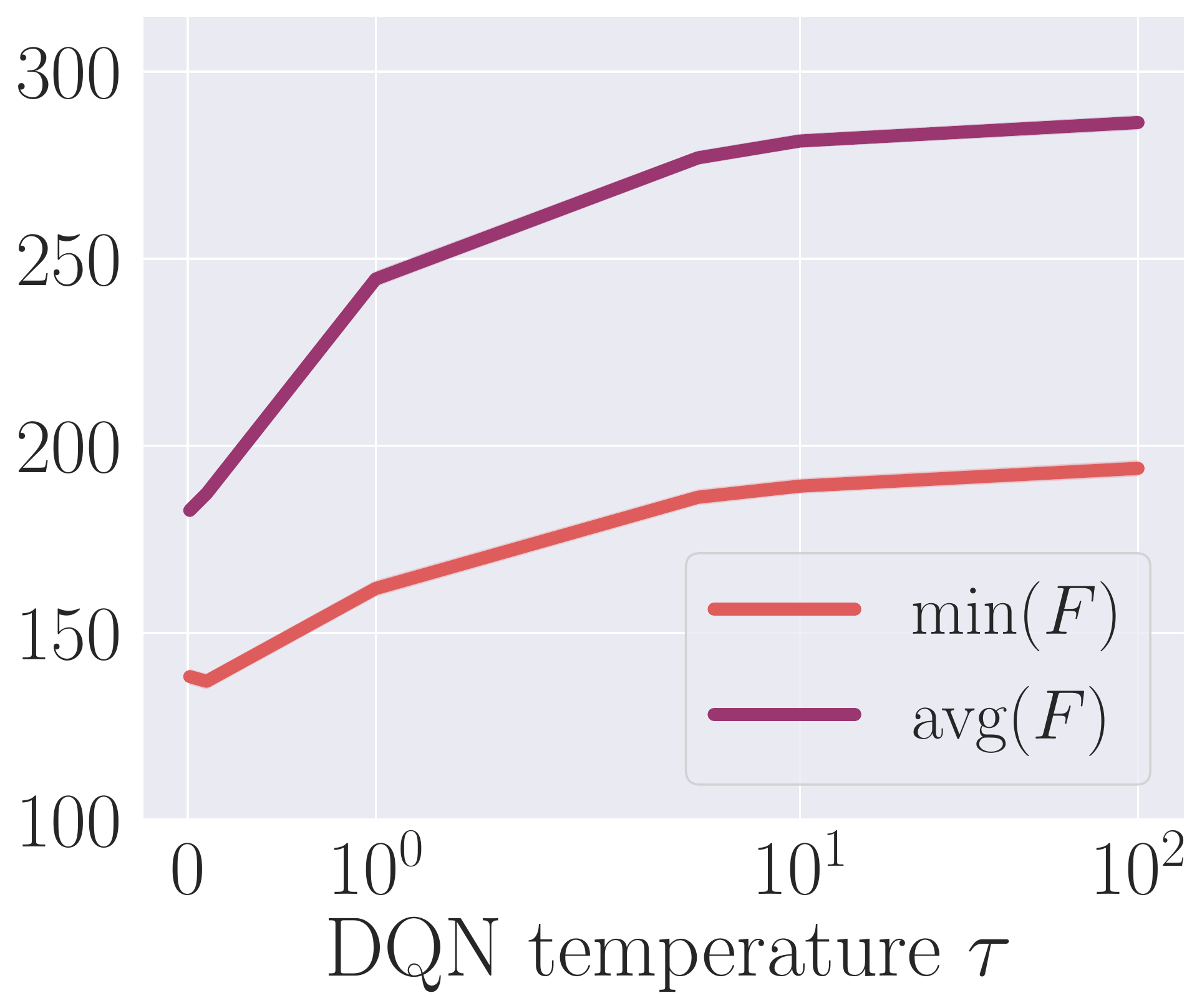}
\vspace{-\baselineskip} 
\caption{Values of $F$ when varying DQN temperature in ALNS. Lower is better.}
\label{fig:temp}
\end{center}
\vspace{-3.25\baselineskip}
\end{wrapfigure}

\vspace{2mm}
\noindent \textbf{Impact of DQN Temperature.} As discussed in Section~\ref{subsec3.3}, the temperature parameter $\tau$ controls the greediness of the resulting policy. Figure~\ref{fig:temp} shows the minimum and mean $F$ obtained with ALNS as a function of $\tau \in \{10^{-2}, 10^{-1}, 10^0, 10^1, 10^2\}$, averaged over the $3$ instance sets. Even though a probabilistic policy may be desirable in some ALNS scenarios, we find that performance generally degrades as the temperature increases. This suggests that, in the settings tested, the inherent stochasticity of the operators is sufficient to explore the search space without the need to combine different choices.

%% file: tables/mdp_evaluation.tex
\begin{tabular}{c ccc c ccc c ccc} %
\toprule
&\multicolumn{3}{c}{\textbf{C-instance}} && \multicolumn{3}{c}{\textbf{R-instance}} && \multicolumn{3}{c}{\textbf{RC-instance}} \\ [-1pt]
\cmidrule(lr){2-4} \cmidrule(lr){6-8} \cmidrule(lr){10-12}
\textbf{$|\mathcal{D}|$} & \textbf{DQN} & \textbf{RAN} & \textbf{LRW} && \textbf{DQN} & \textbf{RAN} & \textbf{LRW} && \textbf{DQN} & \textbf{RAN} & \textbf{LRW}  \\ [-1pt]
\midrule
\textbf{2}         & 232.2±2.9         & \textbf{252.2±3.6}            & 252.1±4.5           && 216.3±5           & \textbf{222.9±3.7}            & 222.8±4.4         && 240.2±7.5         & \textbf{259.4±3.4}            & 258.9±5.6          \\[-2pt]
\textbf{3}         & 228.6±9.0           & 221.7±3.5            & \textbf{245.9±6.0}             && 212.9±5.6         & 208.4±4.7            & \textbf{215.1±3.8}         && \textbf{236.1±6.1}         & 224.0±3.4              & 230.8±7.6          \\[-2pt]
\textbf{4}         & 232.5±5.1         & 221.2±6.4            & \textbf{240.3±5.8}         && \textbf{220.2±3.1}         & 206.9±6.0              & 216.2±4.1         && \textbf{241.2±5.8}         & 222.4±5.1            & 239.5±4.8          \\[-2pt]
\textbf{5}          & \textbf{328.8±2.4}         & 258.5±5.5            & 293.3±8.0          && \textbf{330.9±4.2 }        & 246.8±3.7            & 273.9±6.1         && \textbf{331.1±2.9}         & 260.9±4.1            & 272.5±7.6          \\[-2pt]
\textbf{6}          & \textbf{329.9±4.2}         & 231.7±5.8            & 284.9±8.9        && \textbf{329.5±3.3}         & 217.5±5.5            & 272.5±14.9        && \textbf{329.5±2.6}         & 239.5±5.4            & 261.6±5.1          \\[-2pt]
\textbf{7}          & \textbf{328.5±2.8}         & 246.7±4.4            & 282.4±11.0         && \textbf{330.5±3.8}         & 236.1±4.8            & 253.7±8.1         && \textbf{331.7±2.9}         & 247.6±3.8            & 264.6±6.4          \\[-2pt]
\textbf{8}          & \textbf{329.5±3.9}         & 235.6±5.2            & 281.6±10.6       && \textbf{330.9±3.6}         & 220.4±2.0              & 264.5±7.0           && \textbf{333.7±3.3}         & 243.7±4.5            & 254.7±4.8          \\[-2pt]
\textbf{9}          & \textbf{330.7±3.1}         & 225.8±5.7            & 274.6±12.2       && \textbf{328.9±4.6}         & 212.6±3.8            & 260.3±5.7         && \textbf{332.4±3.7}         & 226.7±7.0              & 250.3±8.4          \\[-2pt]
\textbf{10}         & \textbf{330.2±4.5}         & 224.4±4.7            & 276.2±9.3        && \textbf{330.3±3.3}         & 206.7±4.5            & 258.6±7.0           && \textbf{331.0±2.6}           & 224.3±5.8            & 252.6±15.3         \\[-2pt]
\textbf{11}         & \textbf{330.3±2.9}         & 222.0±4.8              & 275.9±6.4        && \textbf{327.2±8.0}           & 210.4±4.6            & 259.1±9.0           && \textbf{326.1±15.2}        & 223.3±6.6            & 245.7±8.3          \\[-2pt]
\textbf{12}         & \textbf{361.8±0.2}         & 246.9±5.5            & 323.7±7.0          && \textbf{404.2±0.6}         & 246.8±7.0              & 313.1±17.7        && \textbf{354.9±4.1}         & 258.5±4.1            & 283.9±13.0           \\ [-1pt]
\midrule
\textbf{mean} & \textbf{305.7±3.7}         & 235.2±5.0            & 275.5±8.2  && \textbf{305.6±4.6}         & 221.4±4.6            & 255.4±8.0         && \textbf{308.0±5.2}         & 239.1±4.8            & 255.9±7.9          \\
\bottomrule
\end{tabular} 

%% file: tables/alns_evaluation.tex
\begin{tabular}{c c ccccccccccc c} 
\hline
\textbf{C-inst} & \textbf{$|\mathcal{D}|$} & \textbf{2} & \textbf{3} & \textbf{4} & \textbf{5} & \textbf{6} & \textbf{7} & \textbf{8} & \textbf{9} & \textbf{10} & \textbf{11} & \textbf{12} & \textbf{mean}  \\
\hline
\multirow{2}{*}{\textbf{DQN}} &
\textit{avg}  & 316.28          & 314.96          & 311.18          & \textbf{216.33} & \textbf{213.03} & \textbf{216.34} & \textbf{215.12} & \textbf{215.75} & \textbf{214.19} & \textbf{213.09} & \textbf{188.11} & \textbf{239.49}  \\[-3pt]
& \textit{min}  & 244.64          & 245.28          & 240.32          & \textbf{154.06} & \textbf{149.29} & \textbf{153.05} & \textbf{151.87} & \textbf{153.04} & \textbf{150.7}  & \textbf{149.99} & \textbf{146.71} & \textbf{176.27}  \\
\multirow{2}{*}{\textbf{LRW}} &
\textit{avg}  & \textbf{293.84} & \textbf{299.56} & \textbf{302.66} & 248.31          & 259.06          & 261.17          & 266.59          & 264.65          & 265.62          & 264.78          & 224.49          & 268.25           \\[-3pt]
& \textit{min}  & \textbf{209.8}  & \textbf{211.62} & \textbf{207.6}  & 172.97          & 180.96          & 178.07          & 187.71          & 183.88          & 182.07          & 185.3           & 160.95          & 187.36           \\
\multirow{2}{*}{\textbf{RAN}} &
\textit{avg}  & 293.5           & 313.98          & 315.05          & 284.31          & 299.86          & 294.57          & 299.64          & 303.58          & 312             & 305.74          & 286.08          & 300.76           \\[-3pt]
& \textit{min}  & 209.95          & 219.79          & 212.33          & 194.6           & 206.28          & 199.23          & 203.63          & 208.35          & 212.2           & 213.38          & 197.54          & 207.03           \\
\multirow{2}{*}{\textbf{CRW}} &
\textit{avg}  & 293.94          & 314.56          & 314.28          & 285.53          & 299.51          & 295.77          & 302.67          & 307.64          & 311.32          & 308.53          & 286.75          & 301.86           \\[-3pt]
& \textit{min}  & 210.27          & 222.46          & 213.42          & 197.82          & 208.25          & 198.37          & 206.6           & 210.7           & 213.69          & 213.38          & 194.63          & 208.14           \\
\hline
\textbf{R-inst} & \textbf{$|\mathcal{D}|$} & \textbf{2} & \textbf{3} & \textbf{4} & \textbf{5} & \textbf{6} & \textbf{7} & \textbf{8} & \textbf{9} & \textbf{10} & \textbf{11} & \textbf{12} & \textbf{mean}  \\
\hline
\multirow{2}{*}{\textbf{DQN}} &
\textit{avg}  & 330.37          & 334.09          & 331.24          & \textbf{217.25} & \textbf{215.27} & \textbf{215.41} & \textbf{215.89} & \textbf{216.17} & \textbf{215.91} & \textbf{221.71} & \textbf{159.29} & \textbf{242.96}  \\[-3pt]
& \textit{min}  & 273.81          & 266.66          & 272.16          & \textbf{144.97} & \textbf{143.56} & \textbf{144.11} & \textbf{144.19} & \textbf{144.58} & \textbf{144.49} & \textbf{153.64} & \textbf{108.07} & \textbf{176.39}  \\
\multirow{2}{*}{\textbf{LRW}} &
\textit{avg}  & \textbf{322.22} & \textbf{329.14} & \textbf{327.18} & 269.97          & 265.26          & 286.74          & 281.67          & 286.63          & 284.94          & 274.17          & 238.46          & 287.85           \\[-3pt]
& \textit{min}  & \textbf{246.19} & \textbf{254.83} & \textbf{243.38} & 187.33          & 183.51          & 199.16          & 196.51          & 201.83          & 196.82          & 189.54          & 157.33          & 205.13           \\
\multirow{2}{*}{\textbf{RAN}} &
\textit{avg}  & 323.25          & 337.03          & 337.68          & 298.76          & 317.84          & 310.1           & 321.11          & 321.76          & 327.19          & 321.3           & 294.45          & 319.13           \\[-3pt]
& \textit{min}  & 253.64          & 257.11          & 250.05          & 211.7           & 235.07          & 218.61          & 234.45          & 231.79          & 234.31          & 234.38          & 206.54          & 233.42           \\
\multirow{2}{*}{\textbf{CRW}} &
\textit{avg}  & 322.71          & 333.59          & 336.81          & 298.38          & 320.5           & 308.7           & 318.42          & 321.34          & 327.17          & 324.96          & 296.05          & 318.97           \\[-3pt]
& \textit{min}  & 252.39          & 255.69          & 250.28          & 211.47          & 235.15          & 215.55          & 230.06          & 232.49          & 235.87          & 233.28          & 208.18          & 232.76           \\
\hline
\textbf{RC-inst} & \textbf{$|\mathcal{D}|$} & \textbf{2} & \textbf{3} & \textbf{4} & \textbf{5} & \textbf{6} & \textbf{7} & \textbf{8} & \textbf{9} & \textbf{10} & \textbf{11} & \textbf{12} & \textbf{mean}  \\
\hline 
\multirow{2}{*}{\textbf{DQN}} &
\textit{avg}  & 306.06          & 311.02          & 313.78          & \textbf{216.84} & \textbf{218.75} & \textbf{218.16} & \textbf{217.66} & \textbf{217.48} & \textbf{218.12} & \textbf{224.56} & \textbf{201.81} & \textbf{242.2}   \\[-3pt]
& \textit{min}  & 228.16          & 235.04          & 240.93          & \textbf{151.89} & \textbf{154.74} & \textbf{153.71} & \textbf{152.85} & \textbf{152.06} & \textbf{153.79} & \textbf{161.32} & \textbf{159.09} & \textbf{176.69}  \\
\multirow{2}{*}{\textbf{LRW}} &
\textit{avg}  & \textbf{294.94} & \textbf{314.93} & \textbf{310.02} & 278.18          & 282.01          & 286.05          & 292.6           & 293.97          & 293.44          & 292.15          & 261.08          & 290.85           \\[-3pt]
& \textit{min}  & \textbf{205.05} & \textbf{216.86} & \textbf{209.63} & 187.78          & 189.92          & 190.45          & 197.61          & 196.75          & 197.62          & 196.21          & 172.7           & 196.42           \\
\multirow{2}{*}{\textbf{RAN}} &
\textit{avg}  & 297.2           & 319.87          & 317.7           & 288.95          & 298.41          & 294.26          & 300.61          & 308.9           & 315.42          & 305.53          & 283.71          & 302.78           \\[-3pt]
& \textit{min}  & 207.17          & 222.85          & 209.95          & 193.42          & 198.33          & 195.88          & 200.04          & 207.6           & 210.9           & 204.25          & 185.29          & 203.24           \\
\multirow{2}{*}{\textbf{CRW}} &
\textit{avg}  & \textbf{294.94} & 319.07          & 320.76          & 289.51          & 299.99          & 296.09          & 299.98          & 310.67          & 315.4           & 309.69          & 285.73          & 303.8            \\[-3pt]
& \textit{min}  & \textbf{205.05} & 221.27          & 213.01          & 195.39          & 200.23          & 196.03          & 202.8           & 207.05          & 211.47          & 205.59          & 186.24          & 204.01 \\
\hline
\end{tabular}

%% file: sections/5_conclusion.tex
\section{Conclusions and Future Research}
In this work, we have proposed an operator selection mechanism based on Deep Reinforcement Learning to enhance the performance of the ALNS metaheuristic. A key insight and contribution is the proposal of an operator selector that is conditioned on the decision space characteristics of the current solution. We have demonstrated its ability to outperform the classic Roulette Wheel and random operator selection, as well as the potential of using Graph Neural Networks to scale the model to large problem instances. Our results also highlight the impact of the operator portfolio size and the destroy scale on performance. Plans for future work involve applications to other combinatorial optimization problems. 

\vspace{2mm}
\noindent \textbf{Acknowledgements.} This work was partially supported by The Alan Turing Institute under the Enrichment Scheme and the UK EPSRC grant EP/N510129/1.

\vspace{-2mm}